%% file: main.tex
\pdfoutput=1
\documentclass[lettersize,journal]{IEEEtran}
\usepackage[T1]{fontenc}    % use 8-bit T1 fonts
\usepackage{hyperref}       % hyperlinks
\usepackage{url}            % simple URL typesetting
\usepackage{booktabs}       % professional-quality tables
\usepackage{amsfonts}       % blackboard math symbols
\usepackage{nicefrac}       % compact symbols for 1/2, etc.
\usepackage{microtype}      % microtypography
\usepackage{xcolor}         % colors
\usepackage{gensymb}
\usepackage{multirow}
\usepackage{makecell}
\usepackage{array}
\usepackage{graphicx}
\usepackage{amsmath}
\usepackage{float}
\usepackage{subfig}

\usepackage{textcomp}
\usepackage{xspace}
\newcommand{\method}{PSM\xspace}

\usepackage{algorithm}
\usepackage{algorithmic}
\ifCLASSINFOpdf
\else
\fi
\begin{document}
	%
	% paper title
	% Titles are generally capitalized except for words such as a, an, and, as,
	% at, but, by, for, in, nor, of, on, or, the, to and up, which are usually
	% not capitalized unless they are the first or last word of the title.
	% Linebreaks \\ can be used within to get better formatting as desired.
	% Do not put math or special symbols in the title.
	\title{Weakly Supervised Temporal Sentence Grounding via Positive Sample Mining}

	% author names and affiliations
	% transmag papers use the long conference author name format.
	\author{Lu Dong, Haiyu Zhang, Hongjie Zhang, Yifei Huang,   Zhen-Hua Ling,~\IEEEmembership{Senior Member,~IEEE}, Yu Qiao,~\IEEEmembership{Senior Member,~IEEE}, Limin Wang,~\IEEEmembership{Member,~IEEE}, Yali Wang
		% <-this % stops a space
		% \thanks{This paper was supported in part by XXX.}% <-this % stops a space
		\thanks{Lu Dong is with University of Science and Technology of China, Hefei 230027, China, and also with Shanghai Artificial Intelligence Laboratory, Shanghai 202150, China. (Email: dl1111@mail.ustc.edu.cn)}
            \thanks{Haiyu Zhang is with  Beihang University, Beijing 100191, China, and also with Shanghai Artificial Intelligence Laboratory, Shanghai 202150, China. (Email: zhyzhy@buaa.edu.cn)}
            \thanks{Hongjie Zhang, Yifei Huang,  and Yu Qiao are with Shanghai Artificial Intelligence Laboratory, Shanghai 202150, China. (Email: nju.zhanghongjie@gmail.com, hyf015@gmail.com,  and yu.qiao@siat.ac.cn)}
            \thanks{Zhen-Hua Ling is with University of Science and Technology of China, Hefei 230027, China (Email: zhling@ustc.edu.cn)}
            \thanks{Limin Wang is with State Key Laboratory for Novel Software Technology, Nanjing University, Nanjing 210023, China, and also with Shanghai Artificial Intelligence Laboratory, Shanghai 202150, China. (Email: lmwang@nju.edu.cn)}
            \thanks{Yali Wang is with Shenzhen Key Laboratory of Computer Vision and Pattern Recognition, Shenzhen Institute of Advanced Technology, Chinese Academy of Sciences, Shenzhen 518055, China, and also with Shanghai Artificial Intelligence Laboratory, Shanghai 202150, China. (Email: yl.wang@siat.ac.cn).
            \textit{Yali Wang is the corresponding author}.}
            \thanks{Copyright \copyright\ 2025 IEEE. Personal use of this material is permitted. However, permission to use this material for any other purposes must be obtained from the IEEE by sending an email to pubs-permissions@ieee.org.}
	}
	
	% \author{Xinyu Xu, Huazhen Liu, Feiming Wei, Huilin Xiong, \\ Wenxian Yu,~\IEEEmembership{Senior Member,~IEEE}, and Tao Zhang$^{\ast}$ \thanks{*Corresponding author},~\IEEEmembership{Member,~IEEE},
	% 	% <-this % stops a space
	% 	\thanks{This paper was supported in part by XXX.}% <-this % stops a space
	% 	\thanks{X. Xu, T. Zhang, F. Wei, H. Xiong, and W. Yu are with the Shanghai Key Laboratory of Intelligent Sensing and Recognition, School of Sensing Science and Engineering, Shanghai Jiao Tong University, Shanghai 200240, China.}
	% 	\thanks{H. Liu is with the Intelligent Photoelectric Sensing Institute, School of Sensing Science and Engineering, Shanghai Jiao Tong University, Shanghai 200240, China.}
	% }

	% The paper headers
	\markboth{IEEE Transactions on Circuits and Systems for Video Technology}%
	{Shell \MakeLowercase{\textit{et al.}}: Bare Demo of IEEEtran.cls for IEEE Transactions on Magnetics Journals}
	% The only time the second header will appear is for the odd numbered pages
	% after the title page when using the twoside option.
	% 
	% *** Note that you probably will NOT want to include the author's ***
	% *** name in the headers of peer review papers.                   ***
	% You can use \ifCLASSOPTIONpeerreview for conditional compilation here if
	% you desire.

	% If you want to put a publisher's ID mark on the page you can do it like
	% this:
	%\IEEEpubid{0000--0000/00\$00.00~\copyright~2015 IEEE}
	% Remember, if you use this you must call \IEEEpubidadjcol in the second
	% column for its text to clear the IEEEpubid mark.

	% use for special paper notices
	%\IEEEspecialpapernotice{(Invited Paper)}

	% for Transactions on Magnetics papers, we must declare the abstract and
	% index terms PRIOR to the title within the \IEEEtitleabstractindextext
	% IEEEtran command as these need to go into the title area created by
	% \maketitle.
	% As a general rule, do not put math, special symbols or citations
	% in the abstract or keywords.
	\IEEEtitleabstractindextext{%
		\input{chapters/1-abstract}
		
		% Note that keywords are not normally used for peerreview papers.
		}

	% make the title area
	\maketitle

	% To allow for easy dual compilation without having to reenter the
	% abstract/keywords data, the \IEEEtitleabstractindextext text will
	% not be used in maketitle, but will appear (i.e., to be "transported")
	% here as \IEEEdisplaynontitleabstractindextext when the compsoc 
	% or transmag modes are not selected <OR> if conference mode is selected 
	% - because all conference papers position the abstract like regular
	% papers do.
	\IEEEdisplaynontitleabstractindextext
	% \IEEEdisplaynontitleabstractindextext has no effect when using
	% compsoc or transmag under a non-conference mode.

	% For peer review papers, you can put extra information on the cover
	% page as needed:
	% \ifCLASSOPTIONpeerreview
	% \begin{center} \bfseries EDICS Category: 3-BBND \end{center}
	% \fi
	%
	% For peerreview papers, this IEEEtran command inserts a page break and
	% creates the second title. It will be ignored for other modes.
	\IEEEpeerreviewmaketitle

	\input{chapters/2-introduction}

	\input{chapters/3-related_work}

	\input{chapters/4-method}

\input{chapters/5-experiment}
        
        \input{chapters/6-conclusion}

	\input{chapters/9-acknowledgement}
	
	% if have a single appendix:
	%\appendix[Proof of the Zonklar Equations]
	% or
	%\appendix  % for no appendix heading
	% do not use \section anymore after \appendix, only \section*
	% is possibly needed
	
	% use appendices with more than one appendix
	% then use \section to start each appendix
	% you must declare a \section before using any
	% \subsection or using \label (\appendices by itself
	% starts a section numbered zero.)
	%

	% use section* for acknowledgment
	%\section*{Acknowledgment}

	%The authors would like to thank...

	% Can use something like this to put references on a page
	% by themselves when using endfloat and the captionsoff option.
	\ifCLASSOPTIONcaptionsoff
	\newpage
	\fi

	% trigger a \newpage just before the given reference
	% number - used to balance the columns on the last page
	% adjust value as needed - may need to be readjusted if
	% the document is modified later
	%\IEEEtriggeratref{8}
	% The "triggered" command can be changed if desired:
	%\IEEEtriggercmd{\enlargethispage{-5in}}
	
	% references section
	
	% can use a bibliography generated by BibTeX as a .bbl file
	% BibTeX documentation can be easily obtained at:
	% http://mirror.ctan.org/biblio/bibtex/contrib/doc/
	% The IEEEtran BibTeX style support page is at:
	% http://www.michaelshell.org/tex/ieeetran/bibtex/
	%\bibliographystyle{IEEEtran}
	% argument is your BibTeX string definitions and bibliography database(s)
	%\bibliography{IEEEabrv,../bib/paper}
	%
	% <OR> manually copy in the resultant .bbl file
	% set second argument of \begin to the number of references
	% (used to reserve space for the reference number labels box)
	% \begin{thebibliography}{1}
	
	% \end{thebibliography}
	{
		\small
		\bibliographystyle{unsrt}
		\bibliography{ref}
	}

\end{document}

%% file: chapters/1-abstract.tex
\begin{abstract}
			The task of weakly supervised temporal sentence grounding (WSTSG) aims to detect temporal intervals corresponding to a language description from untrimmed videos with only video-level video-language correspondence.
            For an anchor sample, most existing approaches generate negative samples either from other videos or within the same video for contrastive learning. However, some training samples are highly similar to the anchor sample, directly regarding them as negative samples leads to difficulties for optimization and ignores the correlations between these similar samples and the anchor sample.
            To address this, we propose Positive Sample Mining (\method), a novel framework that mines positive samples from the training set to provide more discriminative supervision. 
            Specifically, for a given anchor sample, we partition the remaining training set into semantically similar and dissimilar subsets based on the similarity of their text queries. 
            To effectively leverage these correlations, we introduce a \method-guided contrastive loss to ensure that the anchor proposal is closer to similar samples and further from dissimilar ones. Additionally, we design a \method-guided rank loss to ensure that similar samples are closer to the anchor proposal than to the negative intra-video proposal, aiming to distinguish the anchor proposal and the negative intra-video proposal.
            Experiments on the WSTSG and grounded VideoQA tasks demonstrate the effectiveness and superiority of our method.
    \end{abstract}

%% file: chapters/2-introduction.tex
\section{Introduction}

    \input{figures/motivation}

	\IEEEPARstart{T}{emporal} sentence grounding is a fundamental task for video understanding, where the goal is to locate the start and end timestamps of one segment in the video that semantically corresponds to the given sentence query. This task has the potential for a wide range of applications such as video summarization~\cite{zhang2016video, rochan2018video,huang2024egoexolearn}, video retrieval~\cite{dong2019dual,gabeur2020multi,wang2021t2vlad} and human-computer interaction systems~\cite{huang2020mutual,chen2020look,huang2018predicting}. 
For temporal sentence grounding, the supervised approaches~\cite{gao2017tall,xiao2021boundary,lin2023univtg} have achieved impressive results but heavily rely on manual annotations of temporal locations for each query-video pair. The annotation process is both time-consuming and label-intensive, limiting the scalability of these methods.

To address this limitation,  weakly supervised temporal sentence grounding (WSTSG) approaches~\cite{gao2017tall,ge2019mac,zhang2020learning, zheng2023progressive,zheng2022weakly,kim2024gaussian} have been proposed, requiring only query-video pairs without temporal annotations for training. This reduces the dependency on manual labeling, enabling the use of large-scale data that is readily available online, as query-video pairs can be easily obtained from the Internet. 
Due to the absence of temporal annotations, most previous studies mine negative samples from the training set for contrastive learning, aiming to provide supervision signals for generating proposals. These weakly supervised methods for temporal sentence grounding are broadly categorized into multiple instance learning (MIL)-based and reconstruction-based approaches.
Specifically, MIL-based approaches~\cite{gao2017tall,gao2019wslln,huang2021cross,tan2021logan} generate negative samples from mismatched query-video pairs (\textit{e.g.}, a sentence query paired with a mismatched video in the training set). The model is trained to distinguish between matched (positive) and mismatched (negative) samples. 
On the other hand, reconstruction-based researches~\cite{lin2020weakly,song2020weakly,zheng2022weakly2,zheng2022weakly,kim2024gaussian} generate negative proposals from segments within the same video, enhancing the model’s ability to handle more confusing video segments. 

Given an anchor query-video sample, all previous methods mine negative samples from the training set for contrastive training. However, they treat all samples of  remaining training set equally, ignoring the inherent diversity among them.
As shown in Fig.~\ref{fig:motivation}, for an anchor sample with the query ``Person walking in the room open a refrigerator'', there exists a very similar training sample with the query ``Person open the refrigerator door''. As they have the same object ``refrigerator'' and action ``open'', their corresponding proposal segments are also semantically similar.
If these similar samples are treated as negative samples, the model is forced to push away samples that are semantically alike. This complicates optimization process and leads to suboptimal results.
Therefore, for each anchor sample, the remaining training set should be partitioned into two subsets: a \textit{similar} subset and a \textit{dissimilar} subset, based on the similarity of each sample to the anchor sample. Different operations can then be applied to each subset to better leverage their unique characteristics.

Building on this insight, we propose Positive Sample Mining (\method), a novel WSTSG framework that mines positive samples for each anchor sample, which aims to leverage more discriminative supervision from other video samples.
Specifically, for an anchor sample, we first divide other samples in the remaining training set based on the semantic similarity of their text queries. 
This will result in a similar sample subset and a dissimilar sample subset. 
To discriminate these similar and dissimilar sample pairs, we propose a \method-guided contrastive loss to ensure that the anchor proposal is closer to the similar subset and further from the dissimilar subset, thereby enhancing the model’s ability to capture sophisticated semantic correlations in the sample space.
Furthermore, we propose a \method-guided rank loss to leverage the intra-video information,
which guarantees that similar samples are closer to the anchor proposal than to the negative intra-video proposal.
It can effectively enhance the distinction between the anchor proposal and the negative intra-video proposal.

In summary, our contributions are threefold:
\begin{itemize}
\item We propose a Positive Sample Mining (\method) for the WSTSG task. To the best of our knowledge, this is \textit{the first attempt} to mine positive samples within the training set for this task.
\item We propose a \method-guided contrastive loss to discriminate between similar and dissimilar samples. Furthermore, we design a \method-guided rank loss to enhance the distinction between the anchor proposal and the negative intra-video proposal.% 
\item We validate the effectiveness of our approach through comprehensive evaluations on two tasks:  WSTSG and grounded VideoQA. Our approach achieves superior performance compared to existing methods.
\end{itemize}

%% file: figures/motivation.tex
\begin{figure}[h] % [h] 代表尽量放在当前位置
    \centering
    \includegraphics[width=1.0\linewidth]{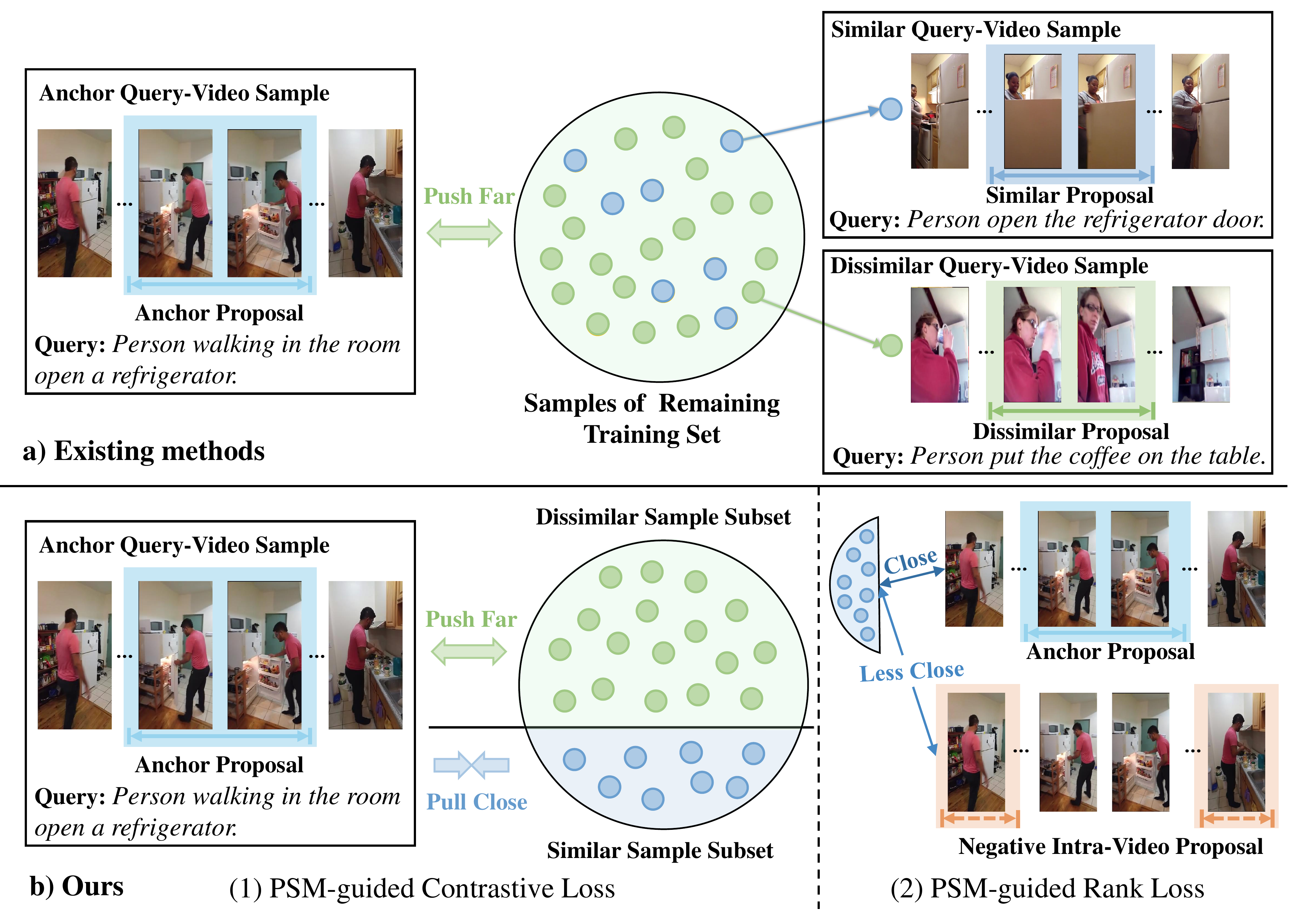} % 调整图片宽度
    \vspace{-4mm}
    \caption{a) Existing methods regard all samples in the remaining training set as negative and push them further from the anchor proposal, which ignores the diversity of the training set. b) Our method mines similar training samples for each anchor sample to leverage more discriminative supervision from other video samples.  Two schemes are proposed. 1)  A \method-guided contrastive loss is proposed to distinguish between similar and dissimilar sample pairs. 2) A \method-guided rank loss is proposed to improve the distinction between the anchor proposal and the negative intra-video proposal.
    }
    \label{fig:motivation} % 为图片设置引用标签
\end{figure}

%% file: chapters/3-related_work.tex
\section{Related Works}

\subsection{Fully Supervised Temporal Sentence Grounding}
Many previous works~\cite{gao2017tall,anne2017localizing,ge2019mac,jiang2019cross,chen2018temporally,yuan2019semantic,zhang2021multi,zhou2022thinking,hu2023camg,sun2023video,zhang2023video,qi2024collaborative} focus on fully supervised temporal sentence grounding, where each query-video pair is annotated with precise start and end timestamps. 
TALL~\cite{gao2017tall} first produces proposals of various lengths through sliding windows, then uses query-proposal fusion for moment regression.
 % Anchor-based
2D-TAN~\cite{zhang2020learning} uses a two-dimensional map to model the temporal relation between proposal candidates,  enabling an exhaustive enumeration of proposals with diverse durations.
CRNet~\cite{sun2023video} formulates temporal sentence grounding as a video reading comprehension task, enabling the model to capture comprehensive relational information from multiple perspectives.
P-Debias~\cite{qi2024collaborative} investigates both visual and combinatorial biases within the training dataset to enhance the robustness of the training process.
However, these methods require manual annotations of temporal locations for each query-video pair. These manual annotations are usually label-consuming and subjective (inconsistency among multiple annotators)~\cite{otani2020uncovering}, which ultimately  harms the scalability of these approaches for real-world applications.

\subsection{Weakly Supervised Temporal Sentence Grounding}
To reduce annotation effects, the weakly supervised paradigm has gained increasing attention across various video tasks, such as temporal action localization~\cite{zhang2021cola,wang2021exploring,hu2023learning,zhang2023cross,zhao2024snippets}, video instance segmentation~\cite{yan2022solve,yinhui2023weakly} and video anomaly detection~\cite{yang2023towards,wu2024vadclip,zhou2024batchnorm}. Regarding temporal sentence grounding (TSG), the weakly supervised TSG methods use only video-level query-video correspondence during training, without requiring start and end timestamps. In the absence of explicit timestamp annotations, most existing approaches adopt the anchor text query and the anchor proposal as positive pairs and mine negative samples for contrastive learning. 
One group of methods adopts the multi-instance learning (MIL) approach~\cite{mithun2019weakly,gao2019wslln,huang2021cross,ma2020vlanet,tan2021logan}. These methods construct negative samples from mismatched query-video pairs (\textit{e.g.}, an anchor video paired with a mismatched query in the training set). The model learns the video-level video-language correspondence by maximizing the matching scores of the positive pairs while penalizing those of the negative pairs.
 Reconstruction-based methods~\cite{lin2020weakly,song2020weakly,zheng2022weakly2,zheng2022weakly,huang2023weakly,kim2024gaussian} measure the query-proposal alignment by the reconstruction of the query, which provides more fine-grained information. Additionally, these methods mine easy and hard negative intra-video proposals to facilitate contrastive learning.

\input{figures/framework}

Both MIL-based and reconstruction-based methods treat all remaining samples in the training set as negative samples equally. However, they ignore the inherent diversity in the training set, where some samples are highly similar to the anchor sample. Therefore, the model sometimes would be forced to push away two samples that are highly similar, leading to difficulties for optimization. Moreover, these methods ignore the correlations between similar samples and the anchor sample. In contrast, our approach identifies the similar and dissimilar samples in the training set and ensures the anchor proposal is closer
to the similar subset and further from the dissimilar subset, which alleviates the difficulty of capturing correlations between query-video samples and leverages
more discriminative supervision from other video samples.

\subsection{Negative Sample Mining}
Negative sample mining~\cite{jin2018unsupervised,robinson2020contrastive,zhang2021video,yoon2022selective,yoon2023counterfactual,lan2023curriculum,qi2024bias} has been widely studied in metrics learning and has massive applications for supervised and weakly supervised TSG. Most of these methods observe that it is helpful to use negative samples that are difficult for the model to discriminate. ~\cite{robinson2020contrastive} theoretically proves hard negative objectives can capture desirable generalization properties for contrastive learning.  In supervised TSG, negative sample mining focuses on debiasing the shortcut of memorizing spurious correlations that link given queries to specific scenes. SQuiDNet~\cite{yoon2022selective} performs selective debiasing via disentangling good and bad retrieval bias based on the query meaning. CTDL~\cite{yoon2023counterfactual} mitigates the bias of TSG training via contrasting factual retrievals with counterfactually biased retrievals. Multi-NA~\cite{lan2023curriculum} proposes multiple negative augmentations to diversify the data distribution, which eases the problem of fitting the temporal distribution bias. BSSARD~\cite{qi2024bias} introduces adversarial training on synthesized bias-conflicting samples to mitigate uneven temporal distributions of target moments. For weakly supervised TSG, current methodologies mine hard intra-video negative proposals for robust training. PPS~\cite{kim2024gaussian} generates diverse negative proposals through multiple Gaussian masks. DSCE~\cite{kim2024learnable} gradually increases the difficulty of the negative proposal with a dual-signed cross-entropy loss. However, most of the existing methods treat the samples of the remaining training set equally, which ignores the inherent diversity among them. In contrast, our \method partitions the remaining training set into similar and dissimilar subsets for each anchor sample and discriminates these similar and dissimilar samples, enhancing the model's ability to capture sophisticated semantic correlations in the sample space and providing more discriminative supervision for weakly supervised TSG.

\subsection{Positive Inter-Sample Mining}
Positive inter-sample mining approaches extract positive signals from other samples in the training set. The goal is to explore latent correlations and interactions between multiple data points. Positive inter-sample mining has various applications in computer vision, including image retrieval~\cite{wu2022contextual,suma2025ames,xie2024d3still}, domain adaptation~\cite{titov2011domain,wang2024exploiting} and video object detection~\cite{han2020mining,qi2024imc}. CSD~\cite{wu2022contextual} mines similar images to establish a contextual similarity constraint on the anchor image, which effectively guides the learning process of lightweight query models. HVR-Net~\cite{han2020mining} identifies similar samples with distinct object labels, utilizing them as hard negative samples to improve the effectiveness of contrastive learning. To the best of our knowledge, we are the first to mine positive inter-video samples for the weakly supervised temporal sentence grounding and demonstrate the effectiveness of our method.

%% file: figures/framework.tex
\begin{figure*}[t]
	\centering
	\includegraphics[width=1.0\textwidth]{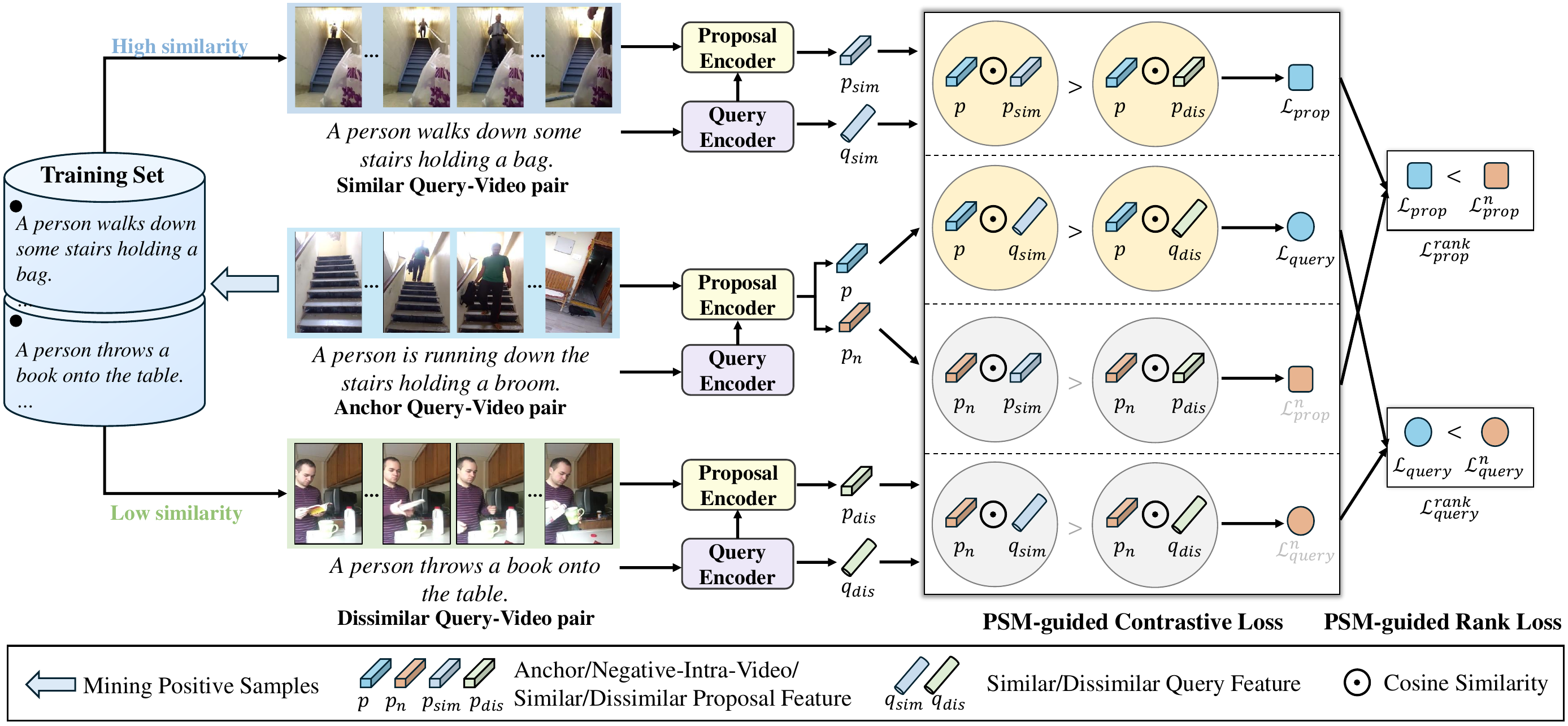} 
	\caption{
    Overview of our proposed method.  First, we mine positive samples from the remaining training set to capture sophisticated semantic correlations in the sample space. Next, the proposal and query encoders are adopted to extract the global representations of proposals and queries. For similar and dissimilar query-video pairs, we utilize the positive proposal feature from the proposal encoder to compute the PSM-guided contrastive loss. Finally, \method-guided contrastive loss is adopted to distinguish between similar and dissimilar queries/proposals for the anchor proposal. Furthermore, a \method-guided rank loss improves the distinction between the anchor proposal and the negative intra-video proposal, based on the distance to similar samples (\textit{i.e.}, the \method guided contrastive loss). 
    }
	\label{fig2:framework}
\end{figure*}

% \begin{figure}[h] % [h] 代表尽量放在当前位置
%     \centering
%     \includegraphics[width=0.8\linewidth]{figures/framework.pdf} % 调整图片宽度
%     \caption{motivation}
%     \label{fig:single_column} % 为图片设置引用标签
% \end{figure}

%% file: chapters/4-method.tex
\section{Method}

\input{figures/query_and_proposal_encoder}
We first present the problem formulation of weakly supervised temporal sentence grounding. Given a video $\mathcal{V}$ and a query sentence $\mathcal{Q}$ that describes a sub-event within this video, the objective is to ground the sentence to a specific temporal segment within the video by predicting the start and end timestamps. During training, the ground truth of the timestamps is not available.

The overall architecture of our \method is illustrated in Fig. \ref{fig2:framework}. \method consists of a proposal encoder, a query encoder,  mining positive samples, and two \method-guided losses. The query and proposal encoders extract the features of the query and the proposal using the transformer-based backbones \cite{zheng2022weakly,kim2024gaussian}. We then pool the query and proposal features to extract their global representations. In addition, we mine
similar and dissimilar query-video sample pairs from the remaining training set to capture sophisticated semantic correlations in the sample space. To distinguish between similar and dissimilar sample pairs, we propose a \method-guided contrastive loss to bring the anchor proposal closer to similar samples and further from dissimilar samples. Furthermore, we propose a \method-guided rank loss to improve the distinction between the anchor proposal and the negative intra-video proposal. The \method-guided rank loss ensures that the similar sample is
closer to the anchor proposal than to the negative intra-video proposal. 

\subsection{Query and Proposal Encoders}

The architectures of query and proposal encoders are illustrated in Fig. \ref{fig:query_and_proposal_encoder}. Given a video and sentence query,  we extract the global proposal and query representations using transformer-based ~\cite{vaswani2017attention} architecture, following previous works~\cite{zheng2022weakly,kim2024gaussian}. For proposal representations, two types are extracted, one is the positive proposal that represents the video segment corresponding to the sentence query, and the other is the negative intra-video proposal that represents the video segment that is not aligned to the sentence query.

Specifically,
given a video and a sentence query, we first use a video backbone to obtain video features and a word embedding to obtain query embeddings. For the video, the pre-trained 3D Convolutional Neural Network~\cite{tran2015learning} encodes segment-level video features. For the sentence query, the pre-trained GloVe~\cite{pennington2014glove} word embeddings are adopted. 
The query encoder adopts a transformer encoder structure~\cite{vaswani2017attention}. To encode the global representation of the query features, we apply mean pooling over all word features to obtain the sentence-level query feature. 
Then, we utilize the proposal encoder, which is based on a transformer, to handle the multi-modal interaction between the video features and word features of the query, following CPL~\cite{zheng2022weakly}.
The proposal encoder outputs the proposal weights for end-to-end training.
To encode the global representation of proposals, we pool the proposal features by computing the weighted sum of raw video features and their corresponding proposal weights.  To discriminate the intra-video information, the proposal encoder encodes two types of proposals: the positive proposal and the negative intra-video proposal.

\subsection{Mining Positive Samples}

For each anchor sample, unlike previous approaches ~\cite{mithun2019weakly,gao2019wslln,huang2021cross,zheng2022weakly2,zheng2022weakly} regarding all remaining samples in the training set as negative, we mine positive samples from the remaining training set to capture sophisticated semantic correlations
in the sample space. Specifically, given a training set with $N$ samples,  for each anchor sample $\mathcal{S}_i$ with sentence query $\mathcal{Q}_i$ and video  $\mathcal{V}_i$, we divide the remaining training set samples into similar and dissimilar sets. we first adopt a pre-trained text encoder $f$, \textit{i.e.}, SentenceTransformer~\cite{reimers-2019-sentence-bert}, to extract the text features of the entire training set.  Then, we compute the cosine similarity between the anchor text query and all remaining text queries in the training set, as 
\begin{equation}
    \text{Mat[i, j]} =\text{cos}\big(f(\mathcal{Q}_i), f(\mathcal{Q}_j)\big). 
\end{equation}
The top-$k$ most similar samples form a similar subset, while the remaining samples form a dissimilar subset, as 
\begin{align}
    \mathcal{S}_{i}^{sim} & = \left\{ (\mathcal{Q}_j, \mathcal{V}_j) \mid j \in \mathop{\text{Top-k-index}}_{j \neq i} \left( \text{Mat[i, j]} \right) \right\}, \\
     \mathcal{S}_{i}^{dis}  &  = \mathcal{S} \setminus \left( \{S_i\} \cup \mathcal{S}_{i}^{sim} \right)
    \text{.}
\end{align}

The process for mining positive samples is outlined in Algorithm~\ref{alg1}. Its runtime consists of two main components: query feature extraction with a complexity of $O(n)$ and the retrieval process with a complexity of $O(n^2)$. In practice,  the retrieval phase demonstrates remarkable efficiency when leveraging  PyTorch's highly-optimized vectorized operations (\textit{i.e.,} topk), typically completing within one second for 10k-level samples. Moreover, as illustrated in Table~\ref{tab:psm_time},  the time required by query feature extraction is usually less than 1\% of the time for weakly supervised training, largely because the feature extraction is a pre-processing step requiring only one epoch, whereas WSTSG training involves multiple epochs. 

Then, we randomly select a similar sample and a dissimilar sample from these two subsets for training, respectively.

\input{tables/algorithm}

Next, we input the anchor, similar, and dissimilar query-video pairs into the query and proposal encoders to extract their global representations, which consist of three parts. The first part is the query representations of similar and dissimilar samples, labeled as  $q_{sim}$ and $q_{dis}$. The second part is the positive proposal representations for the anchor, similar and dissimilar query-video pairs, labeled as $p$,  $p_{sim}$, and $p_{dis}$. The final part is the negative intra-video proposal representation of the anchor sample, labeled as $p_n$. In summary, there are four types of proposal features:  $p$,  $p_{sim}$, $p_{dis}$, and $p_n$, as well as two types of query features: $q_{sim}$ and $q_{dis}$. Then, all query and proposal features are $L_2$ normalized.

\subsection{\method-Guided Contrastive  Loss}
To distinguish between similar and dissimilar sample pairs for each anchor proposal, we propose a \method-guided contrastive loss, which employs a margin loss to ensure that the cosine similarity between the anchor proposal and the similar sample is smaller than that between the anchor proposal and the dissimilar sample. The similar and dissimilar samples can be represented as two modalities: queries or proposals. Both the two-modal versions and the full form of \method-guided contrastive losses are given as
\begin{align}
    \mathcal{L}_{query}  &=
  \mathrm{max}\bigl(p \cdot q_{dis} - p \cdot q_{sim}  +\gamma_1,0\bigr)
  \text{,}
\label{eq:neigh_q}\\
   \mathcal{L}_{prop}  &=
  \mathrm{max}\bigl(p \cdot p_{dis} - p \cdot p_{sim}  +\gamma_2,0\bigr)
  \text{,}
\label{eq:neigh_p}\\
      \mathcal{L}_{\method}^{CL} &=
 \mathcal{L}_{query} + 
 \mathcal{L}_{prop} 
  \text{,}
\label{eq:neigh_q_and_p}
\end{align}
here, the symbol $\cdot$ represents the inner product, which represents the cosine similarity as these features are $L_2$ normalized. $\gamma_1$ and $\gamma_2$ are hyperparameters that control the margin between the cosine similarities of the anchor proposal with respect to the similar and dissimilar samples.

\subsection{\method-Guided Rank Loss}

To distinguish between the anchor proposal and the negative intra-video proposal, we propose a \method-guided rank loss, which adopts a margin loss to ensure that the distance between the anchor proposal and the similar sample is smaller than the distance between the negative intra-video proposal and the similar sample. The similar samples can be represented as two modalities: queries or proposals. We apply the \method-guided contrastive loss to measure the distance between the similar sample and an intra-video proposal, \textit{i.e.}, the anchor proposal or the negative intra-video proposal, because the \method-guided contrastive loss and the distance between the similar sample and an intra-video proposal are positively correlated. That is, a lower \method-guided contrastive loss indicates that the intra-video proposal is closer to similar samples.    
Specifically, the distances between the anchor proposal and similar query $d(p, q_{sim})$ and proposals $d(p, p_{sim})$ can be represented by the \method-guided contrastive losses, which can be represented as
\begin{align}
    &d(p, q_{sim}) = \mathcal{L}_{query}
  \text{,}
\label{eq:equal_p_q_sim}\\
&d(p, p_{sim}) = \mathcal{L}_{prop}
  \text{.}
\label{eq:equal_p_p_sim}
\end{align}

% equal_p_q_sim, d_pn_p
Similarly, the distances between the negative intra-video proposal and similar query $d(p_n, q_{sim})$ and proposals $d(p_n, p_{sim})$ can also be represented by the \method-guided contrastive losses, which are given as
\begin{align}
    d(p_n, q_{sim}) &= \mathcal{L}_{query}^n =
  \mathrm{max}\bigl(p_n \cdot q_{dis} - p_n \cdot q_{sim}  +\gamma_3,0\bigr)
  \text{,}
\label{eq:d_pn_q}\\
d(p_n, p_{sim}) &= \mathcal{L}_{prop}^n  =
  \mathrm{max}\bigl(p_n \cdot p_{dis} - p_n \cdot p_{sim}  +\gamma_4,0\bigr)
  \text{,}
\label{eq:d_pn_p}
\end{align}
here, $\gamma_3$ and $\gamma_4$ are the margin values for the contrastive loss.
Finally, both the two-modal versions and the full form of the \method-guided rank losses are given as

\begin{align}
    \mathcal{L}_{rank}^{query}  &= \mathrm{max}(d(p, q_{sim}) - d(p_n, q_{sim}) + \gamma_5, 0) \notag \\ &=
  \mathrm{max}\bigl(\mathcal{L}_{query} - \mathcal{L}_{query}^{n}  +\gamma_5,0\bigr)
  \text{,}
  \label{eq:rank_q}\\
  \mathcal{L}_{rank}^{prop}  &= \mathrm{max}(d(p, q_{sim}) - d(p_n, q_{sim}) + \gamma_6, 0)  \notag \\ &=
  \mathrm{max}\bigl(\mathcal{L}_{prop} - \mathcal{L}_{prop}^{n}  +\gamma_6,0\bigr)
  \text{,}
  \label{eq:rank_p}\\
  \mathcal{L}_{\method}^{rank} &=
\mathcal{L}_{rank}^{query} + 
\mathcal{L}_{rank}^{prop}
  \text{,}
\label{eq:neigh_q_and_v}
\end{align}
 Where $\gamma_5/\gamma_6$ are hyperparameters that regulate the margin between the distances of the similar sample with respect to the anchor and negative intra-video proposals.

 \subsection{Training and Inference}
During training, we optimize the network using the reconstruction-based loss $\mathcal{L}_{base}$~\cite{zheng2022weakly2,zheng2022weakly,kim2024gaussian} along with our two \method-guided losses $\mathcal{L}_{\method}^{CL}$ and $\mathcal{L}_{\method}^{rank}$,  the overall training loss is given as
\begin{equation}
      \mathcal{L}_{PSM} =
      \mathcal{L}_{base} + 
 \mathcal{L}_{\method}^{CL} +
 \mathcal{L}_{\method}^{rank} 
  \text{.}
\label{eq:final_loss}
\end{equation}

To improve the recall rate of the anchor proposal, we also generate multiple anchor and negative intra-video proposal pairs following ~\cite{zheng2022weakly,kim2024gaussian}. Moreover, the loss is applied only to the anchor proposal with the smallest loss, \textit{i.e.}, the anchor proposal with maximum confidence.

During inference, to select the top-1 proposal from multiple anchor proposals, we adopt the vote-based proposal selection scheme as in previous work~\cite{zheng2022weakly}. To avoid additional inference costs for mining positive samples and computation of two \method-guided losses, we adopt only the reconstruction-based loss $\mathcal{L}_{base}$ for voting among multiple anchor proposals. Therefore, our \method scheme introduces no additional memory and evaluation cost during inference.

%% file: figures/query_and_proposal_encoder.tex
\begin{figure}[t] % [h] 代表尽量放在当前位置, t是居上
    \centering
    \includegraphics[width=1.0\linewidth]{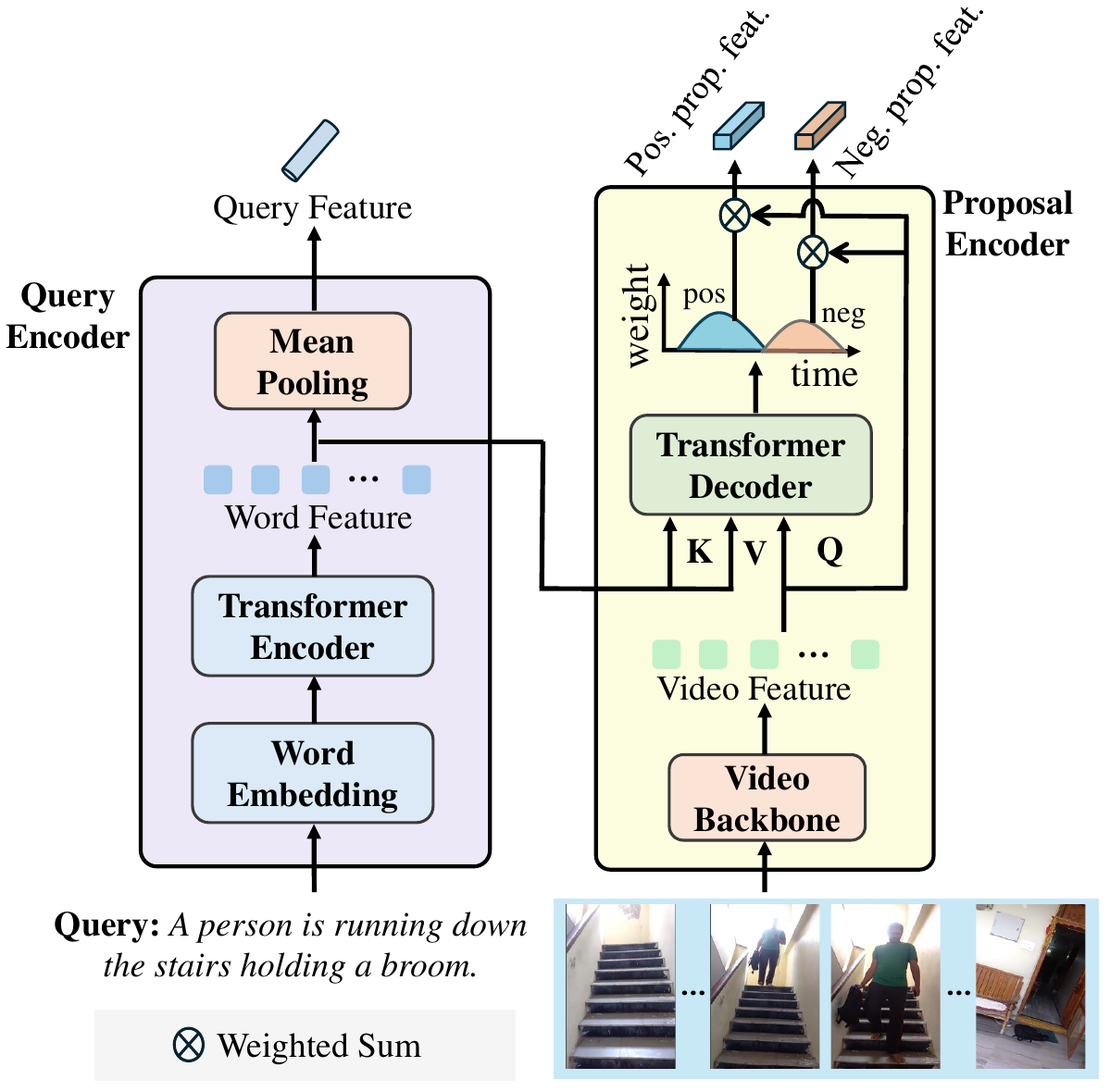} % 调整图片宽度
    \vspace{-4mm}
    \caption{The architecture of query and proposal encoder. The main network is adopted from CPL~\cite{zheng2022weakly}. We add mean-pooling to extract the global query feature and employ weighted sum to extract global positive and negative proposal features.}    
    
    \label{fig:query_and_proposal_encoder} % 为图片设置引用标签
\end{figure}

%% file: tables/algorithm.tex
\begin{algorithm}[t]
    \caption{Mining Positive Samples}
    \label{alg1}
    \begin{algorithmic}[1]
        \STATE \textbf{Input:} $N$ samples, each sample $\mathcal{S}_i$ has a query $\mathcal{Q}_i$ and a video $\mathcal{V}_i$
        \STATE \textbf{Output:} Similar sample dictionary $\mathcal{S}^{sim}$, Dissimilar sample dictionary $\mathcal{S}^{dis}$
        \FOR{$i = 1$ to $N$}
            \STATE Extract query feature $f(\mathcal{Q}_i)$
        \ENDFOR
        % \STATE $L_2$ normalize $f(\mathcal{Q})$
       
        \FOR{$i = 1$ to $N$}
        \FOR{$j = 1$ to $N$}
            \STATE Compute Cosine Similarity $\text{Mat}[i,j]=\text{cos}\big(f(\mathcal{Q}_i), f(\mathcal{Q}_j)\big)$\\
        \ENDFOR
        \STATE Get top-$k$ indices $\text{Index}_i^{sim}=\mathop{\text{Top-k-index}}_{j \neq i} \text{Mat}[i,j]$
            \STATE Get similar sample set \\ $\mathcal{S}_{sim}^{(i)}=\left\{ (\mathcal{Q}_j, \mathcal{V}_j) \mid j \in {\text{Index}}^{sim}_{i} \right\}$
            \STATE Get dissimilar sample set $\mathcal{S}_{dis}^{(i)} = \mathcal{S} \setminus \left\{ \{\mathcal{S}_i\} \cup \mathcal{S}_{sim}^{(i)} \right\}$
        \ENDFOR
        \STATE \textbf{Return} Similar sample dictionary $\mathcal{S}_{sim}$ and Dissimilar sample dictionary $\mathcal{S}_{dis}$
    \end{algorithmic}
\end{algorithm}

%% file: chapters/5-experiment.tex
\section{Experimental Setup}

We conduct extensive experiments to evaluate our \method on two tasks: WSTSG and grounded VideoQA. The WSTSG benchmarks consist of Charades-STA~\cite{gao2017tall} and Activitynet Captions~\cite{krishna2017dense}. The grounded VideoQA benchmark includes NExT-GQA~\cite{xiao2024can}, which consists of two sub-tasks: weakly supervised temporal sentence grounding and keyframe-based question answering tasks. 

\subsection{Datasets}

\begin{itemize}
    \item \textbf{Charades-STA}~\cite{gao2017tall} contains 5.3k/1.3k videos of human daily indoor activities, and 10.6k/3.2k video-query pairs for training/testing. The average video length is 30 seconds. We report our results on the test split. 
    \item \textbf{ActivityNet Captions}~\cite{krishna2017dense} contains 10.0k/4.9k/5.0k videos from YouTube, and  37.0k/17.5k/17.0k video-query pairs for training/validation/testing.  The average video length is 118 seconds. Following
previous methods~\cite{chen2018temporally}, we use $val_2$ as the testing set.
    \item \textbf{NExT-GQA}~\cite{xiao2024can} contains 3.9k/0.6k/1.0k  daily life videos, and 34.1k/3.4k/5.5k question-answering pairs for training/validation/testing. Each QA pair in both validation and test sets is also annotated with an answer interval, while there is no answer interval annotation in the training set. The average video length is 44 seconds. 
    Following~\cite{xiao2024can}, we report our results on the test split.

\end{itemize}

\subsection{Evaluation Metrics}
For the WSTSG dataset,  we use two conventional evaluation metrics introduced in~\cite{gao2017tall}, which are R@$n$,IoU=$m$, and R@$n$,$m$IoU. The R@$n$,IoU=$m$ denotes the percentage of having at least one of the top-$n$ predicted temporal boundaries with temporal Intersection over Union ($t$IoU) larger than the threshold $m$. The R@$n$,$m$IoU is the mean value of the highest $t$IoU in the n predicted temporal boundaries.

For the grounded VideoQA task, we adopt $m$IoU along with three other metrics following~\cite{xiao2024can}, they are (1) $m$IoP: It estimates whether the predicted temporal window lies inside the grounding. (2) Acc@GQA: A new metric proposed in~\cite{xiao2024can}, which denotes the percentages of questions that are both correctly answered and also visually grounded (i.e., IoP$\ge$0.5). (3) Acc@QA: standard metrics for QA task. The R@$n$ is set to  R@1 by default.

\subsection{Implementation Details}
For the WSTSG task, we build  the \method framework based on PPS~\cite{kim2024gaussian}. Following PPS~\cite{kim2024gaussian}, for video segment features, we use C3D features~\cite{tran2015learning} in ActivityNet Captions dataset and I3D features~\cite{carreira2017quo} in Charades-STA dataset. For word embeddings, we adopt GloVe features~\cite{pennington2014glove}.  We set the maximum number of video segments to 200  and the maximum length of the sentence query to 20. For the transformers, the number of transformer layers is 3 and the number of attention heads is 4. The dimension of both visual and textual features is set to 256.  For the masked sentence query, we randomly mask 1/3 of the words. During training, we use the Adam optimizer with a learning rate of 0.0004 and a batch size of 32. For mining positive samples, the top 20 most similar samples are selected to form a similar sample subset. To minimize storage requirements, we store the indices of similar samples for each anchor sample rather than retaining the full data entries. For the \method-guided contrastive loss and \method-guided rank loss hyperparameters, we set them as 
$\gamma_1=\gamma_2=\gamma_3=\gamma_4=0.5$, $\gamma_5=\gamma_6=0.15$.  The grounded VideoQA task includes two sub-tasks: WSTSG and keyframe-based QA. For the WSTSG part, we built our \method framework based on the NG baseline~\cite{xiao2024can}, which adopts a similar Gaussian mask used in CPL~\cite{zheng2022weakly}. For the keyframe-based QA model, following~\cite{xiao2024can}, we adopt two models: Temp[CLIP]~\cite{buch2022revisiting} and FrozenBiLM~\cite{yang2022zero}. We uniformly sample 32 frames for each video. For other training hypermeters, we follow~\cite{xiao2024can}. All $\gamma_i$ ($i=1$ to $6$) hyperparameters adopt the same values as those used in the WSTSG task. For mining positive samples, the top 20 most similar samples form a similar sample subset.

\subsection{Experimental Results}
\input{tables/1-charades_sota}

We evaluate our method against state-of-the-art (SoTA) models on two WSTSG datasets: Charades-STA and ActivityNet Captions, as well as a grounded VideoQA dataset: NExT-GQA. For the WSTSG task, our comparisons include WSTSG models such as WSTAN~\cite{wang2021weakly}, SCANet~\cite{yoon2023scanet} and PPS~\cite{kim2024gaussian}.  The grounded VideoQA task includes two sub-tasks: WSTSG and keyframe-based QA. For the WSTSG sub-task, our comparisons include NG and NG+~\cite{xiao2024can}.  For the keyframe-based QA sub-task, the VideoQA models include Temp~\cite{buch2022revisiting} and FrozenBiLM~\cite{yang2022zero}. Temp model uses multiple features, such as  Swin~\cite{liu2021swin}, CLIP~\cite{radford2021learning} and BLIP~\cite{li2022blip} features. 

\subsubsection{Comparison on Charades-STA and ActivityNet Captions}
Table \ref{tab:charades_sota} and Table \ref{tab:anet_sota} present the comparative results on the Charades-STA and ActivityNet Captions datasets, respectively. The results indicate that our method consistently achieves the best or comparable performance across multiple recalls and IoU thresholds.
Specifically, for Charades-STA dataset,  our method (PPS + \method) outperforms the state-of-the-art PPS~\cite{kim2024gaussian} and SCANet~\cite{yoon2023scanet} methods by margins of 2.83\%, 2.87\%, 2.57\%, 2.03\%, and 2.59\% on R@1,IoU=0.3, R@1,IoU=0.5, R@1,IoU=0.7, R@5,IoU=0.5, and R@5,IoU=0.7, respectively.  For ActivityNet Captions dataset, our method outperforms the state-of-the-art  by 1.52\%, 1.55\%, 1.38\%, and 2.33\% on R@1,IoU=0.5, R@1,IoU=0.7, R@5,IoU=0.3, R@5,IoU=0.5, and R@5,IoU=0.7, respectively. These results highlight the proposals generated by our method has higher quality. The performance boost is primarily achieved by the discriminative supervision from positive samples. 
While UGS~\cite{huang2023weakly} achieves the best performance on R@1,IoU=0.7 for the ActivityNet Captions dataset, its model ensembling strategy relies on multiple reconstruction-based models to aggregate diverse confidence scores during inference, which suffers from reduced diversity in the generated proposals~\cite{huang2023weakly}. This drawback leads to inferior performance on R@1,IoU=0.3 and R@1,IoU=0.5 compared to our method. Moreover, the model ensembling strategy increases computational costs, demanding nearly twice the inference time of our method.

\subsubsection{Comparison on NExT-GQA}
To assess the effectiveness of \method on the grounded VideoQA task, we conduct experiments on the NExT-GQA dataset, with results summarized in Table \ref{tab:nextgqa_sota}. 
The experimental results indicate that our method (NG + \method) achieves significant improvements on the NExT-GQA dataset with mIoP, mIoU, and Acc@GQA.  Specifically, when using the Temp[CLIP]~\cite{buch2022revisiting}  as the VideoQA model, our method outperforms the state-of-the-art NG+~\cite{xiao2024can} method by 2.1\%, 2.7\%, and 1.9\% on mIoP, mIoU, and Acc@GQA, respectively.  Moreover, our method outperforms our baseline NG by 2.0\%, 7.1\%, and 2.4\% on mIoP, mIoU, and Acc@GQA, respectively. When using the FrozenBiLM~\cite{yang2022zero} as the VideoQA model, our method outperforms the state-of-the-art NG+ method by 1.7\%, 3.3\%, and 0.7\% on mIoP, mIoU, and Acc@GQA, respectively. Note that NG+ requires additional annotations of GPT4~\cite{gpt3.5}, whereas our approach achieves state-of-the-art performance without requiring any additional annotations, indicating the superiority of mining positive samples from other video samples. These results signify that our \method effectively guides the VideoQA model to prioritize the correct temporal moments providing visual evidence. Moreover, our method achieves better QA accuracy with more precise visual evidence for both VideoQA baselines, \textit{i.e.}, Temp[CLIP] and FrozenBiLM.

\input{tables/1-anet_sota}
\subsection{Ablation Study}

We conduct extensive ablation experiments on the Charades-STA dataset to explore the effects of different factors, including
different components of PSM, the sample mining strategy, the number of similar query-video pairs, the feature types for mining similar queries,  different backbones, and hyperparameters.
The experimental settings remain consistent with those defined earlier. The findings from these ablation studies provide more in-depth insights into the critical elements that contribute to the effectiveness of our proposed framework.
\input{tables/2-nextgqa_sota}

\input{tables/4_ablation_module}

\subsubsection{Effect of different components of \method}

We use the PPS baseline to ablate the following designs of \method: \method-guided contrastive loss $L_{\method}^{CL}$ and \method-guided rank loss $L_{\method}^{rank}$.  Each has query and proposal modalities, resulting in four losses: $\mathcal{L}_{query}$, $\mathcal{L}_{prop}$, $\mathcal{L}_{query}^{rank}$, and $\mathcal{L}_{prop}^{rank}$.
The results are shown in Table \ref{tab:ablation-module}. Row 1 is the baseline of our model, which does not use positive sample mining.  
 Among the combinations of the four losses, adopting all \method-guided losses yields the best performance. This suggests that leveraging more discriminative supervision from positive samples enables the generated positive proposals to better represent temporal locations. Besides, we conclude some observations based on the results: (1) From the comparison between row 2 to row 5 and row 6 to row 7, we can observe that both the \method-guided contrastive losses $\mathcal{L}_{query}$/$\mathcal{L}_{prop}$ and \method-guided rank losses $\mathcal{L}_{query}^{rank}$/$\mathcal{L}_{prop}^{rank}$ contribute to the improved performance compared to the baseline. Notably, by incorporating intra-video information, which contains richer fine-grained details that are harder for the model to learn, our \method-guided rank losses effectively improve the model's capability and provide a larger boost.
 (2) From the comparison between row 8 to row 9, both the query and the proposal losses have complementary effects, as the two losses provide two modality views for moderate coupling.  (3) From the comparison between row 2 to row 6, row 7 to row 8,  the proposal losses $\mathcal{L}_{prop}$/$\mathcal{L}_{prop}^{rank}$ achieve greater improvements than the query losses $\mathcal{L}_{query}$/$\mathcal{L}_{query}^{rank}$. 
We hypothesize that this phenomenon can be attributed to the video proposal's richer informational content compared to the query, which offers more robust supervision signals for the learning process. These experimental findings collectively demonstrate the significance of each individual component in our proposed framework and highlight their complementary interactions in achieving the best results.

\subsubsection{Effect of sample mining strategy}
\input{tables/ablation_for_other_loss}

% equal_p_q_sim, d_pn_p
We conduct ablation experiments on the Charades-STA dataset to validate our positive sample mining strategy. Specifically, we investigate two variant designs. 1) Hard Negative Mining (HNM): treating the mined similar samples as hard negative instead of positive during training. 2) positive mining with common rank loss (CR loss): we adopt cosine similarity as the distance between the similar sample and an intra-video proposal. Consequently, Equations \ref{eq:equal_p_q_sim}, \ref{eq:equal_p_p_sim}, \ref{eq:d_pn_q} and \ref{eq:d_pn_p} are modified. For example, Equation \ref{eq:equal_p_q_sim} is now defined as $d(p, q_{sim})=p \cdot q_{sim}$.
As shown in Table \ref{tab:ablate_loss_head}, the HNM variant achieves only marginal performance gain (+0.15\% for R@1,mIoU). We attribute this to the inherent design of the PPS baseline:  negative intra-video proposals already serve as effective hard negatives through contrastive learning, making additional negative mining less impactful. For another, the CR loss variant only shows marginal performance over using \method-guided contrastive loss alone (\textit{e.g.,} 46.12\% vs 45.96\% in Table \ref{tab:ablation-module} on R@1,mIoU), and underperforms our method. This result highlights the importance of both constraints to similar and dissimilar samples in \method-guided rank loss, comparing to sole alignment to similar samples in the CR loss
variant.

\input{tables/5_neighbor_number}

\subsubsection{Effect of the number of similar samples}
\label{exp:neighbor_num}
We conduct a comprehensive ablation study to assess the impact of the number of similar samples for positive sample mining on model performance, with the results detailed in Table \ref{tab:ablation_neighbor_num}. We observe the following findings: (1) Using various number of similar samples achieves better results compared to the baseline using no similar sample, even when utilizing only a single similar sample. This highlights the importance of leveraging positive samples as more discriminative supervision signals in the learning process.
(2) As evidenced in the last two rows of Table \ref{tab:ablation_neighbor_num}, the performance gradually saturates as the number of similar samples exceeds a certain threshold. This phenomenon can be explained by the inherent trade-off in mining positive samples: while increasing the number of similar samples can increase diversity, it also inevitably introduces dissimilar samples into the subset of similar samples, which could potentially compromise the model's performance.
\subsubsection{Effect of feature types for mining similar queries}

\input{tables/7-different_q_feat}

The choice of similar samples is critical to the effectiveness of \method, as it directly affects the model's ability to capture sophisticated semantic correlations in the sample space. To investigate this, we choose multiple textual models to retrieve similar queries, including word-level features from GloVe and BERT, as well as sentence-level features from BERT's [CLS] token and SentenceTransformer. For word-level features, we apply mean pooling on the word-level features to obtain the sentence-level representations. Subsequently, all sentence-level features are used to select top-k similar query-video pairs. The final WSTSG results are presented in Table \ref{tab:ablation_different_q_feat}.  Our method with the SentenceTransformer features yields the best results, while our method with the GloVe features performs the worst. We hypothesize that the GloVe features exhibit limited discriminative capability, as evidenced by the cosine similarity matrix of the entire Charades-STA dataset, which ranges narrowly from 0.7 to 1. In contrast, the SentenceTransformer features show a much broader distribution, with cosine similarity values spanning the full range from -1 to 1.

\subsubsection{Effects of different backbones}
To evaluate the generalizability of \method, we build \method 
upon multiple backbones, including CNM~\cite{zheng2022weakly2}, CPL~\cite{zheng2022weakly}, and PPS~\cite{kim2024gaussian}. Table \ref{tab:ablation_different_backbone} reports the results of this ablation study in terms of R@1, IoU=0.5, and mIoU for both R@1 and R@5 metrics.
Specifically, in terms of R@1, IoU=0.5, PSM improves performance by 2.26 for CNM, 3.32 for CPL, and 2.87 for PPS. For R@1, mIoU, PSM leads to gains of 1.89 for CNM, 2.33 for CPL, and 2.05 for PPS.
These results highlight the versatility and effectiveness of our \method in enhancing the performance of different backbones on the WSTSG task. The consistent improvements across all evaluation metrics confirm that our method can be seamlessly integrated into existing networks to achieve better results. The performance of CNM at R@5 is not provided, as CNM only generates one anchor proposal.
As demonstrated in Table \ref{tab:ablation_different_backbone}, our \method exhibits more substantial performance gains when integrated with CPL and PPS compared to CNM. This enhanced performance can be attributed to the inherent advantages of CPL and PPS, which generate multiple anchor proposals and selectively optimize the proposal with the highest confidence score. This multi-proposal approach inherently provides greater robustness than the single-anchor-proposal framework CNM.

\input{tables/3_different_backbone_grounding}

\subsubsection{Effect of hyperparameters}
\input{figures/hyper_ablation}

We conduct comprehensive experiments to evaluate the impact of different combinations of hyperparameters for the \method-guided contrastive loss and the \method-guided rank loss. Specifically, we evaluate $\gamma_1, \gamma_2, \gamma_3$ and $\gamma_4$ for the \method-guided contrastive loss and $\gamma_5$ and $\gamma_6$ for the \method-guided rank loss. When varying each $\gamma_i$,
the remaining $\gamma$ values are fixed. The results are illustrated in Fig. \ref{fig:ablation_hypers}. Through this ablation study, we observe several important findings: (1) the combination of margins ($\gamma_1=\gamma_2=\gamma_3=\gamma_4=0.5, \gamma_5=\gamma_6=0.15$) results in the optimal performance. (2) The model is less sensitive to the variations in $\gamma_1,\gamma_2,\gamma_3,\gamma_4$ than to changes in $\gamma_5,\gamma_6$. (3) The performance significantly degrades when $\gamma_5$ and $\gamma_6$ deviate from their optimal values, with both extremely small and large values. We hypothesize that when these values are too small, the model struggles to distinguish between the anchor proposal and the negative intra-video proposal, which is crucial for WSTSG~\cite{zheng2022weakly}. On the other hand, when $\gamma_5$ and $\gamma_6$ are too large, 
we observe that the model is difficult to optimize in the early training stage, as the initial predictions are much more noisy.

\subsection{Qualitative Analysis}

% figure

\input{figures/topk_neighbor}

\input{figures/case_study}

To validate that our \method effectively captures the correlations between the anchor samples and similar samples, we visualize the encoded proposal space using the Charades-STA dataset. Specifically, two anchor samples are randomly selected and their corresponding top 20 most similar samples are identified. Then, we employ t-SNE~\cite{van2008visualizing} to visualize the proposal features encoded by PPS~\cite{kim2024gaussian} and our method (PPS + \method). From Fig. \ref{fig:topk_neighbor}, we observe that the similar proposal features encoded by PPS are sparsely distributed, and fail to form a clear clustering around the anchor proposal feature. This indicates that PPS is ineffective in grouping proposals with semantics similar to the anchor proposal.  
In contrast, the similar proposal features encoded by our method are tightly clustered around the anchor proposal feature, highlighting our \method's capability to preserve the semantic similarity within the proposal space.
However, for certain query-video pairs, the video clips corresponding to the queries may exhibit complex semantics, where multiple events occur simultaneously. This complexity poses challenges in accurately establishing precise proposals. As a result, some similar proposal features encoded by our method may deviate from the anchor proposal feature.

Fig. \ref{fig:case} shows the qualitative comparison of the results generated by two baselines and our method (PPS + \method). From this figure, we make the following observations: (1) From Fig. \ref{fig:case}(a) (b), and (c), it is notable that  our method can achieve better results than these baselines, proving that our \method technique can capture more accurate query-relevant temporal locations.  (2) As shown in Fig. \ref{fig:case}(a), (b), and (c), CPL and PPS tend to produce longer proposals, while our method can remove the irrelevant proposals near the ground-truth proposals. 
This is mainly because our method mines similar samples from the training set and requires maintaining correlations between the anchor sample and the similar samples. As the proposal length increases, maintaining these correlations becomes more challenging due to the inclusion of irrelevant temporal context, which complicates the feature alignment process.
(3) Fig. \ref{fig:case}(d) shows that the compared methods and our method perform poorly when the query is highly complex. This performance limitation can be attributed to the inherent challenge of identifying sufficiently similar queries in the training set for the complex anchor queries, resulting in weakened discriminative supervision signals from the remaining training samples, ultimately hindering the model's capacity to capture sophisticated correlations between query-video samples.

\input{tables/time_analysis}

We also give a training cost distribution for our method (PPS + \method) on an A100 GPU. The result is shown in Table \ref{tab:psm_time}. We can observe that the feature extraction time and retrieval time of mining positive samples are minimal when compared to the WSTSG training time.  The retrieval phase demonstrates remarkable efficiency since we leverage  PyTorch's highly-optimized vectorized operations (\textit{i.e.,} topk). On the other hand, the feature extraction speed is fast because
the feature extraction is a pre-processing step requiring only
one epoch, whereas WSTSG training involves multiple epochs.

%% file: tables/1-charades_sota.tex
\renewcommand{\arraystretch}{1.2}
\begin{table}[t!]

\centering
\resizebox{\linewidth}{!}{
% \vspace{-0.1in}
    \begin{tabular}{l|ccc|ccc}
 \Xhline{1.0pt}
     \multirow{2}{*}{\makecell[c]{Methods}} & \multicolumn{3}{c|}{R@1,IoU=} & \multicolumn{3}{c}{R@5,IoU=} \\ 
& 0.3 & 0.5 & 0.7 & 0.3 & 0.5 & 0.7 \\ 
\Xhline{0.7pt}

Random          & 20.12 & 8.61 & 3.39 & 68.42 & 37.57 & 14.98 \\
CTF~\cite{chen2020look}         & 39.80 & 27.30 & 12.90 & -     & -     & -     \\
SCN~\cite{lin2020weakly}       & 42.96 & 23.58 & 9.97  & 95.56 & 71.80 & 38.87 \\
WSTAN~\cite{wang2021weakly}     & 43.39 & 29.35 & 12.28 & 93.04 & 76.13 & 41.53 \\
BAR~\cite{wu2020reinforcement}        & 44.97 & 27.04 & 12.23 & -     & -     & -     \\
MARN~\cite{song2020weakly}       & 48.55 & 31.94 & 14.81 & 90.70 & 70.00 & 37.40 \\
CCL~\cite{zhang2020counterfactual}        & -     & 33.21 & 15.68 & -     & 73.50 & 41.87 \\
RTBPN~\cite{zhang2020regularized}    & 60.04 & 32.36 & 13.24 & 97.48 & 71.85 & 41.18 \\
LoGAN~\cite{tan2021logan}      & 51.67 & 34.68 & 14.54 & 92.74 & 74.30 & 39.11 \\
CRM~\cite{huang2021cross}        & 53.66 & 34.76 &16.37 &-     &-     &-     \\
VCA~\cite{wang2021visual}      & 58.58 & 38.13 & 19.57 & 98.08 & 78.75 & 37.75 \\
LCNet~\cite{yang2021local}      & 59.60 & 39.19 & 18.87 & 94.78 & 80.56 & 45.24 \\
CWSTG~\cite{chen2022explore}     & 43.31 & 31.02 & 16.53 & 95.54 & 77.53 & 41.91 \\

CPL~\cite{zheng2022weakly}    & 66.40 & 49.24 & 22.39 & 96.99 & 84.71 & 52.37 \\
CPI~\cite{kong2023dynamic}          & 67.64 & 50.47 & 24.38 & 97.18 & 85.66 & 52.98 \\
CCR~\cite{lv2023counterfactual}        & 68.59 & 50.79 & 23.75 & 96.85 & 84.48 & 52.44 \\

SCANet~\cite{yoon2023scanet}    & 68.04 & 50.85 & 24.07 & 98.24 & \underbar{86.32} & \underbar{53.28} \\

UGS~\cite{huang2023weakly}  & 69.16 & 52.18 & 23.94 & -     & -     & -     \\
OmniD~\cite{bao2024omnipotent}  & 68.30 & 52.31 &24.35 & -     & -     & -     \\
MMDist~\cite{bao2024local}      & 68.90 & 53.29 & 25.27 & -     & -     & -     \\
% \Xhline{0.5pt}
PPS~\cite{kim2024gaussian}        & \underbar{69.06} & \underbar{51.49} & \underbar{26.16} & \underbar{99.18} & 86.23 & 53.01 \\
Ours (PPS + \method)           & \bf 71.89 & \bf  54.36 & \bf 28.73 & \bf 99.19 & \bf 88.35 & \bf 55.87 \\
 \Xhline{1.0pt}
    \end{tabular}
  
    }
      \caption{Performance comparisons on Charades-STA dataset. Bold and underlined numbers denote the best results and the second-best results, respectively. 
      } 
\label{tab:charades_sota}
\end{table}

%% file: tables/1-anet_sota.tex
\renewcommand{\arraystretch}{1.2}
\begin{table}[t!]

\centering
\resizebox{\linewidth}{!}{
% \vspace{-0.1in}
    \begin{tabular}{l|ccc|ccc}
 \Xhline{1.0pt}
     \multirow{2}{*}{\makecell[c]{Methods}} & \multicolumn{3}{c|}{R@1,IoU=} & \multicolumn{3}{c}{R@5,IoU=} \\ 
% & 0.3 & 0.5 & 0.7 & 0.3 & 0.5 & 0.7 \\ 
& 0.1 & 0.3 & 0.5 & 0.1 & 0.3 & 0.5\\ 
\Xhline{0.7pt}

Random          & 38.23 & 18.64 & 7.63  & 75.74 & 52.78 & 29.49 \\
CTF~\cite{chen2020look}         & 74.20 & 44.30 & 23.60 & -     & -     & -     \\
SCN~\cite{lin2020weakly}       & 71.48 & 47.23 & 29.22 & 90.88 & 71.56 & 55.69 \\
WSTAN~\cite{wang2021weakly}     & 79.78 & 52.45 & 30.01 & 93.15 & 79.38 & 63.42 \\
BAR~\cite{wu2020reinforcement}        & -     & 49.03 & 30.73 & -     & -     & -     \\
MARN~\cite{song2020weakly}       & -     & 47.01 & 29.95 & -     & 72.02 & 57.49 \\
CCL~\cite{zhang2020counterfactual}        & -     & 50.12 & 31.07 & -     & 77.36 & 61.29 \\
RTBPN~\cite{zhang2020regularized}    & 73.73 & 49.77 & 29.63 & 93.89 & 79.89 & 60.56 \\
% LoGAN \cite{tan2021logan}      & -     & -     & -     & -     & -     & -     \\
VCA~\cite{wang2021visual}      & 67.96 & 50.45 & 31.00 & 92.14 & 71.79 & 53.83 \\
LCNet~\cite{yang2021local}      & 78.58 & 48.49 & 26.33 & 93.95 & 82.51 & 62.66 \\
CWSTG~\cite{chen2022explore}     & 71.86 & 46.62 & 29.52 & 93.75 & 80.92 & 66.61 \\
CPL~\cite{zheng2022weakly}    & 82.55 & 55.73 & 31.37 & 87.24 & 63.05 & 43.13 \\
% CPL   \cite{zheng2022weakly}    & 82.55 & 55.73 & 31.37 & 87.24 & 63.05 & 43.13 \\
CCR~\cite{lv2023counterfactual}        & 80.32 & 53.21 & 30.39 & 91.44 & 71.97 & 56.50 \\
SCANet~\cite{yoon2023scanet}    & \bf 83.62 & 56.07 & 31.52 & 94.36 & 82.34 & 64.09 \\

CRM~\cite{huang2021cross}        & 81.61 &55.26 & 32.19 & -     &-     & -     \\
OmniD~\cite{bao2024omnipotent}  & 83.24 & 57.34 & 31.60 & -     & -     & -     \\
PPS~\cite{kim2024gaussian}        & 81.84 & \underbar{59.29} & 31.25 & \underbar{95.28} & \underbar{85.54} & \underbar{71.32} \\

MMDist~\cite{bao2024local}      & 83.11 &58.69 & 32.52 &-     &-     &-     \\
UGS~\cite{huang2023weakly}  & 82.10 & 58.07 & \bf 36.91 & -     & -     & -     \\
% \Xhline{0.5pt}
Ours (PPS + \method)            & \underbar{83.25} & \bf 60.81 & \underbar{33.64} & \bf 96.83 & \bf 86.92 & \bf 73.65 \\

 \Xhline{1.0pt}
    \end{tabular}
    }
    \caption{Performance comparisons on ActivityNet Captions dataset. Bold and underlined numbers denote the best results and the second-best results, respectively.
    } 
\label{tab:anet_sota}
\end{table}

%% file: tables/2-nextgqa_sota.tex
\renewcommand{\arraystretch}{1.2}
\begin{table*}[t!]

\centering

\resizebox{0.95\linewidth}{!}{
% \vspace{-0.1in}

    % \begin{tabular}{l|lcc |  ccl | llllllll}
    \begin{tabular}{l|ccc |  ccc | ccc |ccc}
 \Xhline{1.0pt}

WSTSG  & \multirow{2}{*}{QA Model}  & \multirow{2}{*}{Vis Enc} & \multirow{2}{*}{Text Enc}   & \multicolumn{3}{c|}{Acc} & \multicolumn{3}{c|}{IoP} & \multicolumn{3}{c}{IoU} \\
Method &  &  &  &QA &QA$^{\circ}$ & GQA & mIoP & 0.3 & 0.5 & mIoU& 0.3 & 0.5 \\
\Xhline{0.7pt}
Human & -& - & -  & 93.3 & - & 82.1 & 72.1 & 91.7 & 86.2 & 61.2 & 86.9 & 70.3   \\
    Random& - & - & - & 20.0 & 20.0 & 1.7 & 21.1 & 20.6 & 8.7 & 21.1 & 20.6 & 8.7   \\
    \Xhline{0.5pt}
    \multirow{9}*{PH~\cite{xiao2024can}} 
    & VGT~\cite{xiao2022video} & RCNN~\cite{girshick2014rich} & BT~\cite{devlin2018bert}  & 50.9 & 53.8 &12.7 & 24.7 & 26.0 & 24.6 & 3.0 & 4.2 & 1.4  \\
    & VIOLETv2~\cite{fu2023empirical}  & VSWT~\cite{liu2022video} & BT &  52.9 & 57.2 & 12.8 & 23.6 & 25.1 & 23.3 & 3.1 & 4.3 & 1.3  \\
    & VGT  & RCNN & RBT~\cite{liu2019roberta} &  55.7 & 57.7 & 14.4 & 25.3 & 26.4 & 25.3 & 3.0 & 3.6 & 1.7  \\
    & Temp[Swin]~\cite{buch2022revisiting}  & SWT~\cite{liu2021swin} & RBT &  55.9 & 58.7 & 13.5 & 23.1 & 24.7 & 23.0 & 4.9 & 6.6 & 2.3  \\
    & Temp[CLIP]  & ViT-B~\cite{dosovitskiy2020image} & RBT & 57.9 & 60.7 & 14.7 & 24.1 & 26.2 & 24.1 & 6.1 & 8.3 & 3.7  \\
    & Temp[BLIP]  & ViT-B & RBT & 58.5 & 61.5 & 14.9 & 25.0 & 27.8 & 25.3 & 6.9 & 10.0 & 4.5  \\
    & Temp[CLIP] & ViT-L & RBT & 59.4 & 62.5 & 15.2 & 25.4 & 28.2 & 25.5 & 6.6 & 9.3 & 4.1  \\
    & FrozenBiLM~\cite{yang2022zero}  & ViT-L & DBT~\cite{he2020deberta} & 69.1 & 71.8 & 15.8 & 22.7 & 25.8 & 22.1 & 7.1 & 10.0 & 4.4  \\
    \Xhline{0.5pt}
    \multirow{2}*{NG~\cite{xiao2024can}} 
    & Temp[CLIP]  & ViT-L & RBT & 59.4 & 62.7 & 15.5 & 25.8 & 28.8 & 25.9 & 7.7 & 10.9 & 4.6  \\
    & FrozenBiLM & ViT-L & DBT & 70.4 & 73.1 & 17.2 & 24.0 & 28.5 & 23.5 & 9.2 & 13.0 & 5.8  \\
    \Xhline{0.5pt}
    \multirow{2}*{NG+~\cite{xiao2024can}} 
    & Temp[CLIP] & ViT-L & RBT & 60.2  & 63.3  & 16.0 & 25.7  & 31.4 & 25.5 & 12.1 & 17.5 & 8.9   \\
    & FrozenBiLM & ViT-L & DBT & 70.8  & 73.1 & 17.5 & 24.2  & 28.5  & 23.7  & 9.6  & 13.5 & 6.1  \\

    \Xhline{0.5pt}

    % Ours
      \multirow{2}*{Ours (NG + \method)}
    & Temp[CLIP]  & ViT-L & RBT &  61.8  & 65.1  & 17.9 & \bf 27.8  & \bf  34.6 & \bf  26.7 &  \bf 14.8 &  \bf 20.5 & \bf  9.7   \\
    & FrozenBiLM & ViT-L & DBT & \bf 72.1  & \bf 74.5 & \bf 18.2 & 25.9  & 31.4  & 24.5  & 12.9  & 16.4 & 7.7  \\

 \Xhline{1.0pt}
    \end{tabular}
    }
    \caption{Grounded QA performance on the NExT-GQA test set. $\circ$: original NExT-QA.  BT: BERT. RBT: RoBERTa. DBT: DeBERTa-V2-XL. FT5: Flan-T5-XL. Random: always return the entire video duration as the grounding result.} 
\label{tab:nextgqa_sota}
\end{table*}

%% file: tables/4_ablation_module.tex
\renewcommand{\arraystretch}{1.3}
\begin{table}[t!]

\centering

\large
\resizebox{\linewidth}{!}{
% \vspace{-0.1in}

    \begin{tabular}{cc| cc | cc|cc}
 \Xhline{1.0pt}
     \multicolumn{4}{c|}{Loss}  & \multicolumn{2}{c|}{R@1,IoU=} & \multicolumn{2}{c}{R@5,IoU=} \\
    \cline{1-4}
   $\mathcal{L}_{query}$ & $\mathcal{L}_{prop}$   & $\mathcal{L}_{query}^{rank}$ & $\mathcal{L}_{prop}^{rank}$  & 0.5 & mIoU & 0.5 & mIoU \\
\Xhline{0.7pt}
$\times$ & $\times$ & $\times$ & $\times$ &  51.49 & 44.87 & 86.23 & 69.32 \\
    
    \Xhline{0.5pt}
    % single-----
     $\checkmark$ & $\times$ & $\times$ & $\times$   & 52.01 & 45.39 & 87.09 & 69.79 \\
     $\times$ & $\checkmark$ & $\times$ & $\times$  & 52.45 & 45.82 & 87.03 & 69.93 \\
     $\times$ & $\times$ & $\checkmark$ & $\times$  & 52.64 & 46.02 & 87.04 & 70.02 \\
     $\times$ & $\times$ & $\times$ & $\checkmark$  & 52.86 & 46.34 & 87.39 & 70.18 \\
     \Xhline{0.5pt}
     $\checkmark$  & $\checkmark$ & $\times$ & $\times$  & 52.93 & 45.96 & 87.21 & 70.07 \\
     $\times$  & $\times$ & $\checkmark$ &  $\checkmark$  & 53.47 & 46.33 & 87.68 & 70.52 \\
     $\checkmark$ & $\times$ & $\checkmark$ & $\times$  & 53.26 & 46.25 & 87.34 & 70.46 \\
     $\times$ & $\checkmark$ & $\times$ &  $\checkmark$  & 53.81 & 46.56 & 87.77 & 70.88 \\
     \Xhline{0.5pt}
     $\checkmark$ & $\checkmark$ & $\checkmark$ & $\checkmark$  & \textbf{54.36} & \textbf{46.92} & \textbf{88.35} & \bf 71.05 \\
 \Xhline{1.0pt}
    \end{tabular}
    }
    \caption{Ablation study of different components for positive sample mining on the Charades-STA dataset.} 
\label{tab:ablation-module}
\end{table}

%% file: tables/ablation_for_other_loss.tex
\renewcommand{\arraystretch}{1.2}
\begin{table}[t!]

\centering
\resizebox{\linewidth}{!}{
% \vspace{-0.1in}
    \begin{tabular}{c| cc|cc}
 \Xhline{1.0pt}
     \multirow{2} * {Sample Mining Strategy}  & \multicolumn{2}{c|}{R@1,IoU=} & \multicolumn{2}{c}{R@5,IoU=} \\ 
     & 0.5 & mIoU & 0.5 & mIoU \\
\Xhline{0.7pt}
None & 51.49 & 44.87 & 86.23 & 69.32 \\
Hard Negative Mining & 51.85 & 45.02 & 86.72 & 69.49 \\
Positive Mining (CR loss) & 53.47 & 46.12 & 87.43 & 70.49 \\
Positive Mining (Ours) & \bf 54.36 & \bf 46.92 & \bf 88.35 & \bf 71.05 \\
 \Xhline{1.0pt}
    \end{tabular}
    }
    \caption{Ablation study of  sample  mining strategy. The experiment is conducted on the Charades-STA dataset.} 
\label{tab:ablate_loss_head}
\end{table}

%% file: tables/5_neighbor_number.tex
\renewcommand{\arraystretch}{1.2}
\begin{table}[t!]

\centering
\resizebox{\linewidth}{!}{
% \vspace{-0.1in}
    \begin{tabular}{c| cc|cc}
 \Xhline{1.0pt}
    \multirow{2} * {Similar Samples Number} & \multicolumn{2}{c|}{R@1,IoU=} & \multicolumn{2}{c}{R@5,IoU=} \\ 
     & 0.5 & mIoU & 0.5 & mIoU \\
\Xhline{0.7pt}
0 (baseline) & 51.49 & 44.87 & 86.23 & 69.32 \\
    1 & 52.78 & 45.69 & 87.48 & 70.43 \\
    3 & 53.41 & 46.06 & 87.96 & 70.76 \\
    5 & 53.88 & 46.49 & 88.03 & 70.65 \\
    10 & 54.20 & 46.88 & 88.29 & 70.96 \\ 
   20 & \bf54.36 & 46.92 & \bf88.35 &\bf 71.05 \\
   30 & 54.32 & \bf46.95 & 88.31 &\bf 71.05 \\
 \Xhline{1.0pt}
    \end{tabular}
    }
    \caption{Ablation study of the number of similar samples for positive sample mining on the Charades-STA dataset.} 
\label{tab:ablation_neighbor_num}
\end{table}

%% file: tables/7-different_q_feat.tex
\renewcommand{\arraystretch}{1.2}
\begin{table}[t!]

\centering
\resizebox{\linewidth}{!}{
% \vspace{-0.1in}
    \begin{tabular}{c| cc|cc}
 \Xhline{1.0pt}
     \multirow{2} * {Similar Query Feat.}  & \multicolumn{2}{c|}{R@1,IoU=} & \multicolumn{2}{c}{R@5,IoU=} \\ 
     & 0.5 & mIoU & 0.5 & mIoU \\
\Xhline{0.7pt}
Glove [mean] & 53.35 & 46.54 & 87.32 & 70.43 \\
    BERT [mean] & 53.97 & 46.80 & 87.88 & 70.69 \\
    BERT [CLS] & 53.80 & 46.67 & 87.45 & 70.75 \\ 
    SentenceTransformer &  \bf 54.36 &  \bf 46.92 &  \bf 88.35 &  \bf  71.05 \\
 \Xhline{1.0pt}
    \end{tabular}
    }
    \caption{Ablation study of the feature types for mining similar queries in positive sample mining. The experiment is conducted on the Charades-STA dataset.} 
\label{tab:ablation_different_q_feat}
\end{table}

%% file: tables/3_different_backbone_grounding.tex
\renewcommand{\arraystretch}{1.2}
\begin{table}[t!]

\centering
\resizebox{\linewidth}{!}{
% \vspace{-0.1in}
    \begin{tabular}{c|c|cc|cc}
 \Xhline{1.0pt}
    \multirow{2}{*}{Method} & \multirow{2}{*}{+ \method } & \multicolumn{2}{c|}{R@1, IoU=} & \multicolumn{2}{c}{R@5,IoU=} \\
% \cline{3-6}
& & 0.5 & mIoU & 0.5 & mIoU \\
\Xhline{0.7pt}

% 这里填写你的具体数据
\multirow{2} * {CNM~\cite{zheng2022weakly2}}
& $\times$ & 35.43 & 39.54 & - & - \\
& $\checkmark$ & \bf 37.69 & \bf 41.43 & - & - \\

\Xhline{0.5pt}
% 这里填写你的具体数据
\multirow{2} * {CPL~\cite{zheng2022weakly}}
& $\times$ & 49.24 & 42.91 & 84.71 & 67.66 \\
& $\checkmark$ & \bf 52.56 & \bf 45.24 & \bf 86.92 & \bf 69.09 \\

\Xhline{0.5pt}
% 这里填写你的具体数据
\multirow{2} * {PPS~\cite{kim2024gaussian}}
& $\times$ & 51.49 & 44.87 & 86.23 & 69.32 \\
& $\checkmark$ & \bf 54.36 & \bf 46.92 & \bf 88.35 & 
 \bf 71.05 \\

 \Xhline{1.0pt}
    \end{tabular}
    }
    \caption{Results on the Charades-STA datasets when our \method is applied on different backbone networks. Our \method significantly boosts the performance of the existing networks.} 
\label{tab:ablation_different_backbone}
\end{table}

%% file: figures/hyper_ablation.tex
\begin{figure}[t!]
  \centering
  % textwidth or columnwidth
    \includegraphics[width=1\columnwidth]{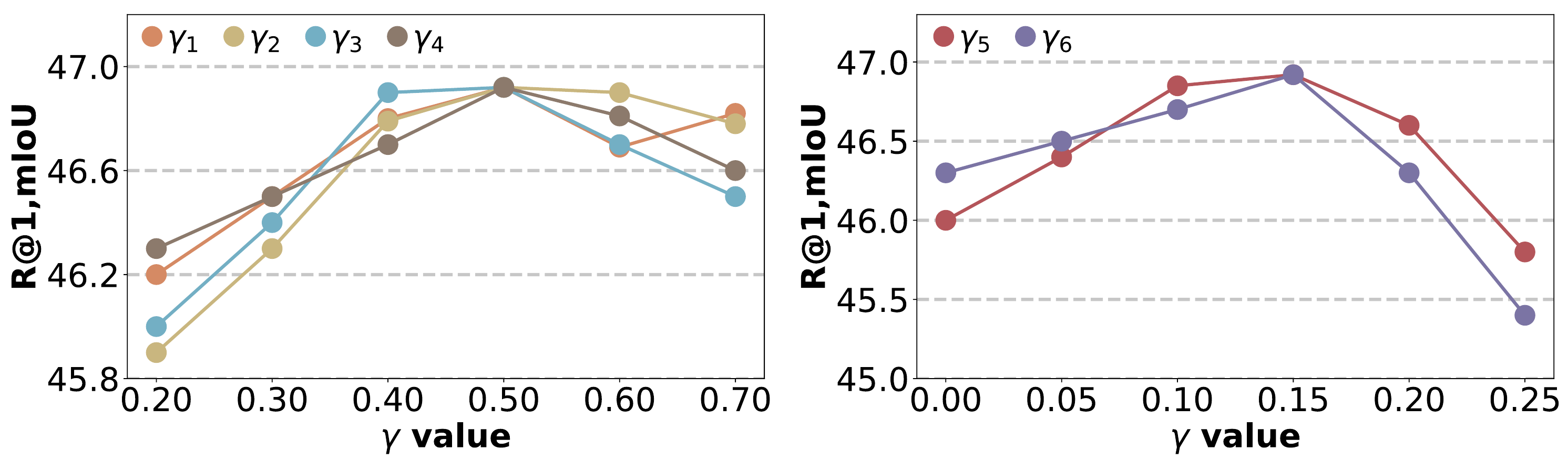}
    \caption{Ablation study by varying $\gamma_1$ to $\gamma_6$ for positive sample mining on the Charades-STA dataset. When varying each $\gamma_i$, the remaining $\gamma$ values are fixed.}
    
    \label{fig:ablation_hypers}
\end{figure}

%% file: figures/topk_neighbor.tex
\begin{figure}[t]
	\centering
	\includegraphics[width=1\columnwidth]{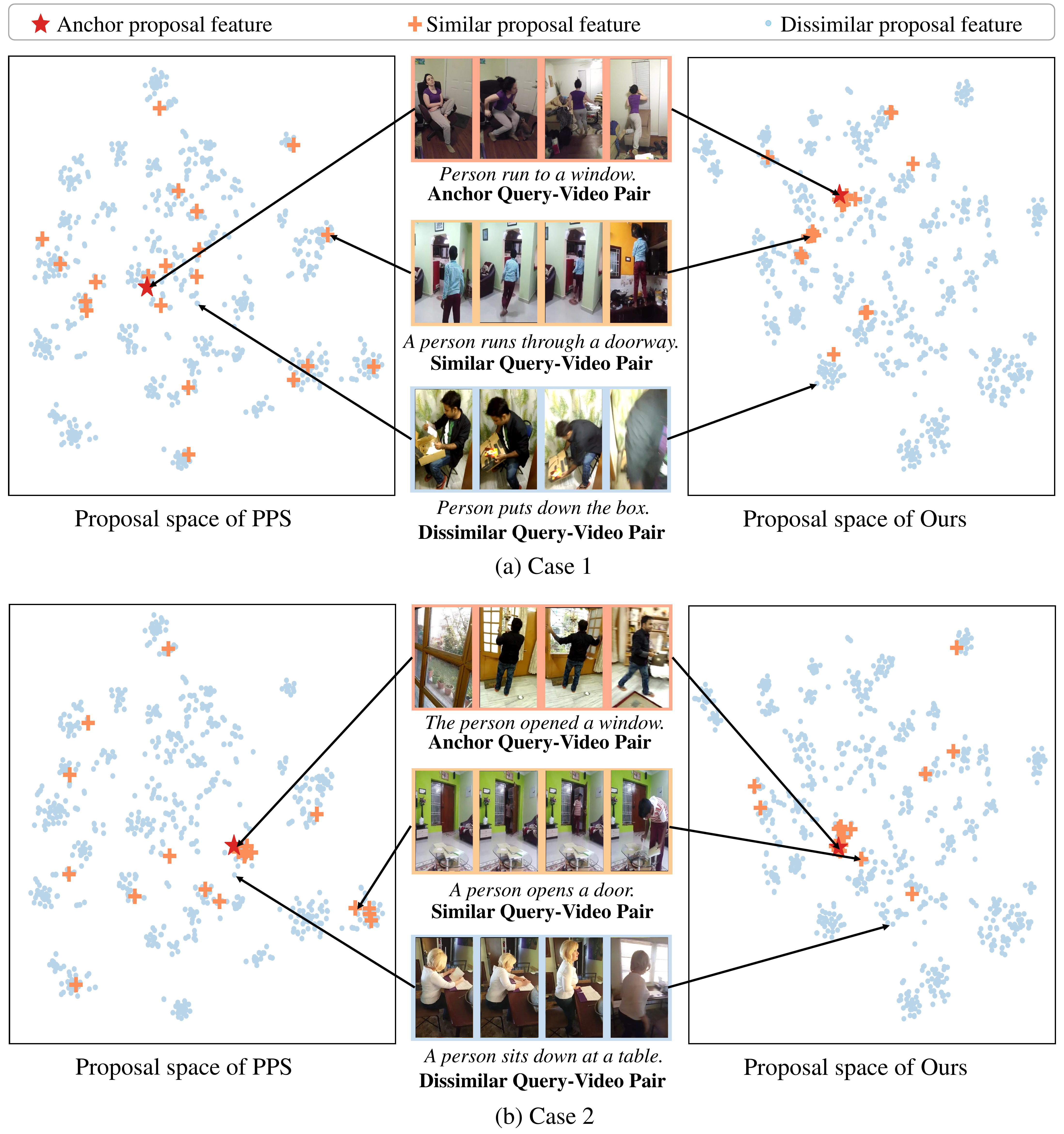} 
	\caption{The t-SNE results of the proposal spaces encoded by baseline PPS and our method (PPS + \method) on the Charades-STA dataset. Two test samples are randomly selected.  For each anchor sample, we present one similar and one dissimilar sample for visualization.} 
	\label{fig:topk_neighbor}
\end{figure}

%% file: figures/case_study.tex
\begin{figure}[t]
	\centering
	\includegraphics[width=1\columnwidth]{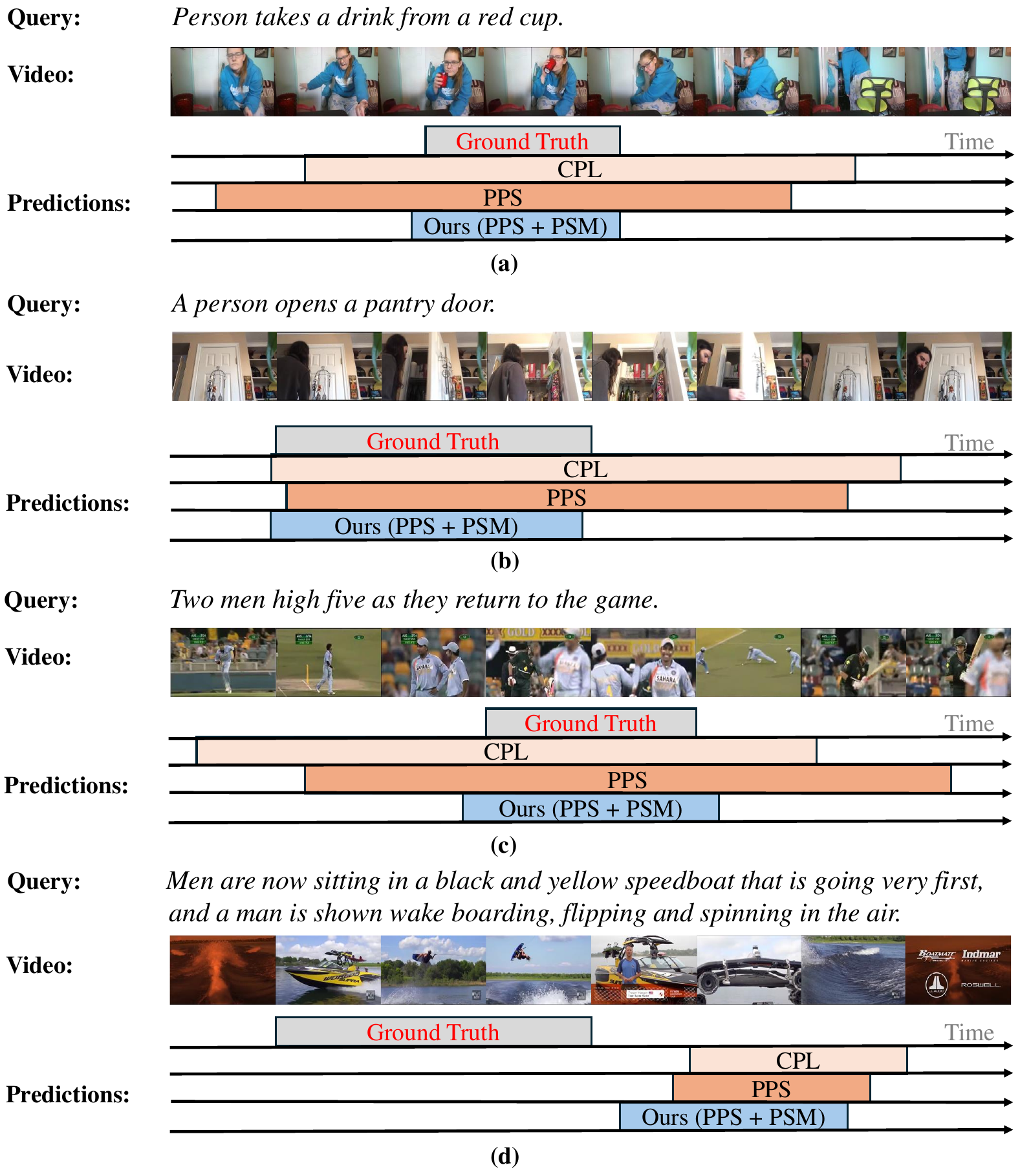} 
	\caption{Qualitative examples of the ground truth, the baselines CPL \cite{zheng2022weakly} and PPS \cite{kim2024gaussian}, as well as our method (PPS + \method). Examples (a, b) are from the Charades-STA dataset, and (c, d) are from the ActivityNet Captions dataset. }
	\label{fig:case}
\end{figure}

%% file: tables/time_analysis.tex
\renewcommand{\arraystretch}{1.2}
\begin{table}[t!]

\centering
\resizebox{\linewidth}{!}{
% \vspace{-0.1in}

% 坐标轴，baseline训练时间,baseline测试时间，our mining positive sample 时间, our training time, our 
    \begin{tabular}{l|c|c}
     \Xhline{1.0pt}
     Dataset & Charades-STA & ActivityNet Captions \\
     \Xhline{0.7pt}
     Training Size & 10.6k & 37.4k \\
     \Xhline{0.7pt}
     Feature Extraction Time (MPS) & 3s & 13s \\
     Retrieval Time (MPS) & 0.07s & 0.58s \\
     WSTSG Training Time & 1867s & 7601s \\
     Total Time & 1870s & 7615s \\
     \Xhline{1.0pt}
    \end{tabular}
    }
    \caption{Training cost distribution for positive sample mining. Here MPS represents mining positive samples. The experiment is conducted on an A100 GPU.} 
\label{tab:psm_time}
\end{table}

% \renewcommand{\arraystretch}{1.2}
% \begin{table}[t!]

% \setlength\tabcolsep{5pt}

% \centering
% \resizebox{\linewidth}{!}{
% % \vspace{-0.1in}

% % 坐标轴，baseline训练时间,baseline测试时间，our mining positive sample 时间, our training time, our 
%     \begin{tabular}{c|c| cc|c}
%  \Xhline{1.0pt}
%     Dataset & Training Size & Mining Time & Training Time  & Extra Time Ratio\\
% \Xhline{0.7pt}
%     Charades-STA & 10.6k & 3s & 30min & 0.2\% \\
%     Activitynet Captions & 37.4k & 13s & 127min & 0.2\% \\
%  \Xhline{1.0pt}
%     \end{tabular}
%     }
%     \caption{Additional time cost for mining positive sample. The experiment is conducted on a A100 GPU.} 
% \label{tab:psm_time}
% \end{table}

%% file: chapters/6-conclusion.tex
\section{Conclusions and future directions}

In this work, we propose the first positive-sample-based method for WSTSG. We mine positive samples for each anchor sample and leverage them as more discriminative supervision. Moreover, we propose a \method-guided contrastive loss to discriminate similar and dissimilar samples. Additionally, we design a \method-guided rank loss to improve the distinction between the anchor proposal and the negative intra-video proposal. Qualitative and quantitative experiments on two tasks, WSTSG and grounded VideoQA, demonstrate the superior performance of \method. Extensive ablation studies further validate the effectiveness of \method. 

While \method demonstrates substantial improvements in the WSTSG task, it has some limitations.  First, \method requires additional training costs for mining similar samples and loss computation. 
Secondly, when the video clips corresponding to the queries contain complex semantics, where multiple events occur simultaneously, it is challenging for our model to output precise proposals. 
Thirdly, our method mines similar samples based on the global semantics of the sentence query, which can be difficult to find similar samples for complex queries. This limitation could potentially be addressed through the decomposition of complex queries into distinct sub-events, and employing a more granular 
 retrieval process that operates on these individual sub-event components, rather than attempting to match entire queries.

%% file: chapters/9-acknowledgement.tex
\section{Acknowledgements}
This work was supported by the National Key R\&D Program of China(NO.2022ZD0160505) and Industry Collaboration Projects Grant, Shanghai Committee of Science and Technology, China (Grant No.22YF1461500).